\documentclass{article}

\PassOptionsToPackage{authoryear, round}{natbib}
 \usepackage[preprint]{neurips_2026}

\usepackage[utf8]{inputenc} 
\usepackage[T1]{fontenc}    
\usepackage[breaklinks=true,bookmarks=true,colorlinks,pagebackref=true]{hyperref}
\usepackage{url}            
\usepackage{booktabs}       
\usepackage{amsfonts}       
\usepackage{nicefrac}       
\usepackage{microtype}      
\usepackage[table]{xcolor}  
\usepackage{marvosym}

\usepackage{graphicx}
\usepackage{amsmath,amssymb}
\usepackage{subcaption}
\usepackage{wrapfig}
\usepackage{multirow}
\usepackage{pifont}
\usepackage{array}
\usepackage{makecell}
\usepackage{adjustbox}
\usepackage{placeins}
\usepackage{float}
\usepackage{wrapfig}

\graphicspath{{figures/}{figures/images/}{NeurIPS26/figures/}{./}}

\def\eg{\emph{e.g.}}

\definecolor{BlueGreen}{HTML}{008080}
\definecolor{OliveGreen}{RGB}{128, 128, 0}

\newcolumntype{I}{!{\vrule width 1pt}}
\makeatletter
\newcommand{\thickhline}{%
    \noalign {\ifnum 0=`}\fi \hrule height 1pt
    \futurelet \reserved@a \@xhline
}
\makeatother

\newcommand{\method}{\mbox{HS\textsuperscript{2}Occ}}

\definecolor{mygreen}{RGB}{93,173,85}
\definecolor{myred}{RGB}{180,0,0}

\definecolor{free}{RGB}{0,0,0}
\definecolor{floor}{RGB}{0,90,200}
\definecolor{partitions}{RGB}{0,170,0}
\definecolor{door}{RGB}{100,80,20}
\definecolor{chair}{RGB}{255,127,80}
\definecolor{table}{RGB}{0,210,255}
\definecolor{sofa}{RGB}{170,150,100}
\definecolor{bed}{RGB}{120,220,0}
\definecolor{appliance}{RGB}{255,140,0}
\definecolor{cabinet}{RGB}{140,30,190}
\definecolor{carpet}{RGB}{255,105,180}
\definecolor{plant}{RGB}{200,0,0}
\definecolor{objects}{RGB}{230,255,60}
\definecolor{suspended_object}{RGB}{120,150,255}

\title{Humanoid-OmniOcc: Stereo-Based Full-View Occupancy Dataset for Embodied AI}

\author{
    Xianda Guo\textsuperscript{1,$^*$\footnotemark[2]} ~~ Bohao Zhang\textsuperscript{2,$^*$} ~~ Chenwei Huang\textsuperscript{2,$^*$} ~~ Shiyuan Chen\textsuperscript{2,3} ~~ Ruilin Wang\textsuperscript{2} \\ \textbf{Yiqun Duan\textsuperscript{4}} ~~ \textbf{Cong Yang\textsuperscript{3}} ~~ \textbf{Qin Zou\textsuperscript{1,\Letter}} ~~ \textbf{Wei Sui\textsuperscript{2,\Letter}} \\
    \textsuperscript{1} School of Computer Science, Wuhan University
    \textsuperscript{2} D-Robotics
    ~\textsuperscript{3} Soochow University ~\textsuperscript{4} UTS\\
    xianda\_guo@163.com, \{bohao.zhang, chenwei.huang, wei.sui\}@d-robotics.cc\\
    $^*$ Equal Contributions \quad\quad \textsuperscript{\Letter} Corresponding Author
\vspace{-5mm}
}

\begin{document}

\maketitle

\renewcommand{\thefootnote}{\fnsymbol{footnote}}
\footnotetext[2]{Work was done during an internship at D-Robotics.}

\begin{abstract}

Occupancy prediction at voxel-level granularity is essential for safe robotic navigation and interaction in complex environments.
Existing occupancy datasets, however, are predominantly designed for autonomous driving with vehicle-centric biases---forward-facing cameras, far-field geometry, and static road priors---limiting their applicability to embodied humanoid perception.
We present \textbf{Humanoid-OmniOcc}, a large-scale panoramic stereo-based occupancy dataset tailored for humanoid robots.
The dataset encompasses 15 diverse simulated indoor scenes and 5 real-world environments, yielding over 155K samples with broad scene and style diversity.
Importantly, the dataset is designed around a \textbf{Real2Sim2Real} closed-loop paradigm: real sensor specifications drive physically accurate simulation, simulation produces large-scale annotated training data, and models trained in simulation are directly evaluated on real-world captures---enabling iterative refinement of the sim-to-real pipeline.
We further propose \textbf{H}umanoid \textbf{S}urround \textbf{S}tereo-guided \textbf{Occ}upancy model (\textbf{\method} ) that exploits robust depth priors for accurate 2D-to-3D lifting.
Extensive experiments show that \method{} consistently outperforms monocular baselines and generalizes well to both unseen simulated test scenes and real-world environments, validating the effectiveness of the Real2Sim2Real design.
Code and data will be available upon acceptance at \url{https://d-robotics-ai-lab.github.io/humanoid-omniocc}.

\end{abstract}
\section{Introduction}

Reliable near-field 3D occupancy perception is essential for humanoid robots to navigate, manipulate, and interact safely within human-centered spaces. Among scene representations, occupancy grids strike a practical balance between fidelity and computability by estimating the free/occupied state of each voxel. This geometry-centric abstraction supports collision-aware planning, integrates naturally with semantics, and scales to real-time control loops.

Most existing 3D occupancy datasets target \emph{autonomous driving}~\cite{semantickitti,surroundocc,openoccupancy}, emphasizing forward-facing views, long-range geometry, and largely static road scenes---assumptions that misalign with humanoid operation. Humanoids work \emph{indoors}, with \emph{egocentric} viewpoints, frequent occlusions, and \emph{dynamic} human activity; consequently, existing benchmarks underspecify the near-field, omnidirectional conditions where safety-critical interaction occurs.

\begin{figure}[t]
    \centering
    \includegraphics[width=\linewidth]{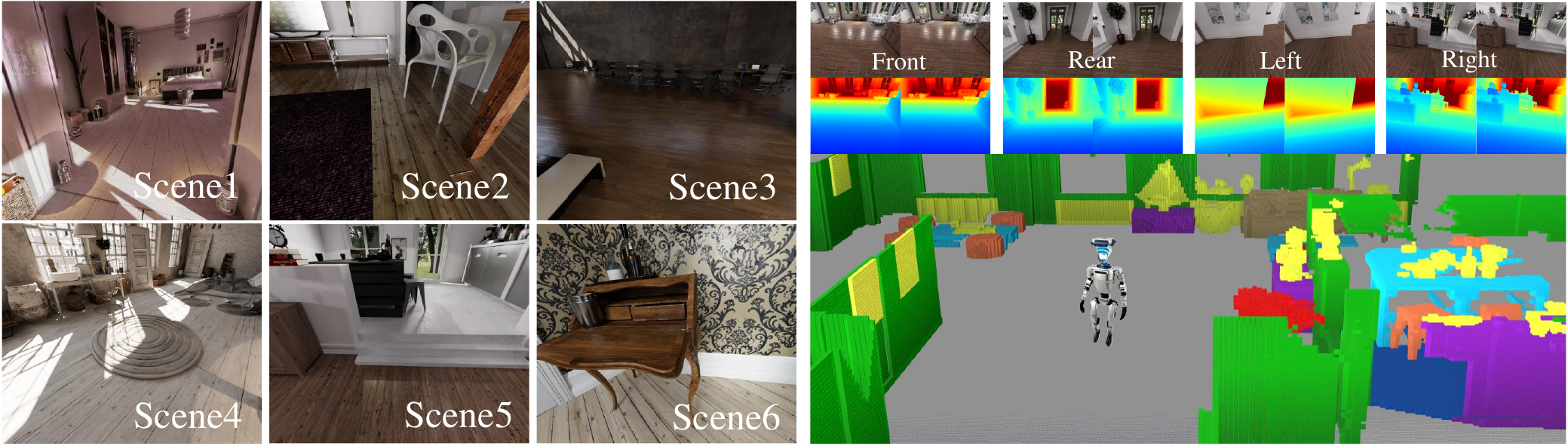}
    \caption{Illustration of the proposed \textbf{Humanoid-OmniOcc} dataset. \textbf{Left:} Six representative scenes rendered in high photorealistic quality, covering diverse spatial layouts and material textures. \textbf{Right:} Visualization of one scene with four stereo RGB pairs (\textit{Front, Rear, Left, Right}), their corresponding depth maps, and voxelized occupancy labels.}
    \label{fig:teaser}
\end{figure}

\begin{wrapfigure}{r}{0.5\linewidth}
    \centering
    \includegraphics[width=0.88\linewidth]{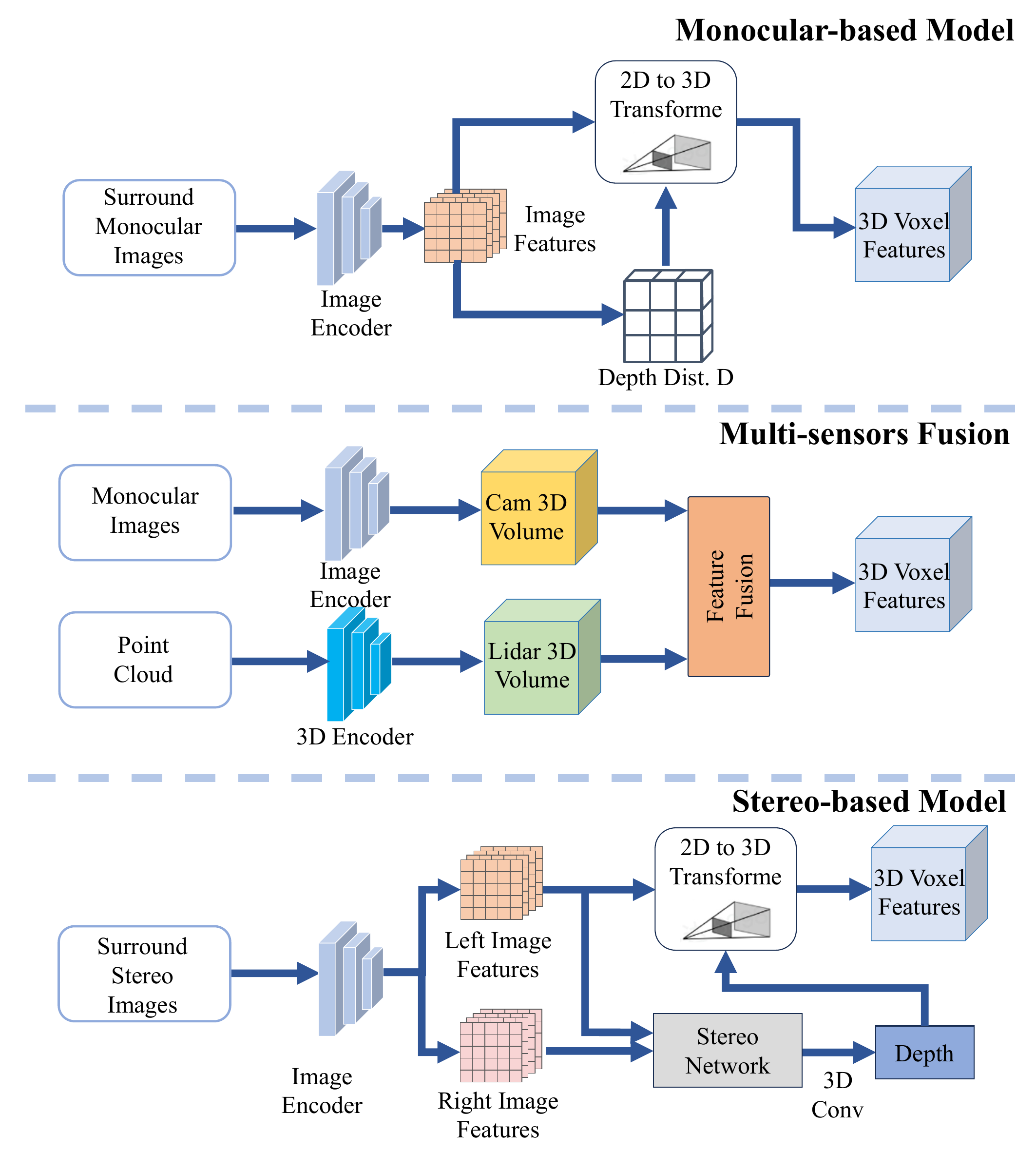}
    \caption{Comparison of different 3D perception paradigms. Top: Monocular-based model. Middle: Multi-sensor fusion model. Bottom: Our proposed stereo-based model.}
    \label{fig:architecture_comparison}
\end{wrapfigure}

More recently, \emph{embodied/indoor} occupancy datasets have begun to emerge (\eg, HumanoidOcc~\cite{humanoidocc}), yet they are predominantly \emph{monocular}---prone to depth ambiguity and weak cross-domain generalization---or rely on \emph{LiDAR}, which is accurate but costly and cumbersome for head-mounted humanoid platforms. As illustrated in Fig.~\ref{fig:architecture_comparison}, these existing perception paradigms either struggle with geometric reliability or incur prohibitive hardware costs. To break this bottleneck at both the data and algorithmic levels, we advocate for a stereo-centric approach. For near-field perception, \emph{stereo} offers a pragmatic middle ground: metrically grounded dense depth at arm's reach with lower cost and easier deployment.
Meanwhile, stereo depth estimation~\cite{guo2023openstereo,stereoanything,guo2025lightstereo} has matured to a degree that makes it a strong source of depth priors.
However, to our knowledge, \emph{no} surround-stereo occupancy dataset exists that aligns with humanoid robots' $360^\circ$ egocentric sensing in indoor environments.

We introduce \textbf{Humanoid-OmniOcc}, a \textit{stereo-based panoramic occupancy} dataset tailored for humanoid perception, designed around a \textbf{Real2Sim2Real} closed-loop paradigm. Built on NVIDIA Isaac Sim, a head-like rig of \textit{four synchronized stereo cameras} provides full-surround RGB.
The Real2Sim2Real pipeline operates as follows: (i)~\textit{Real$\to$Sim}---physical sensor parameters (intrinsics, extrinsics, FoV, baseline) from the Unitree~G1 robot are faithfully replicated in the simulator with photorealistic PBR and sensor-accurate camera models (Fig.~\ref{fig:sensor}); (ii)~\textit{Sim}---large-scale training data with centimeter-level occupancy ground truth is generated via multi-view geometric verification across 15 globally diverse indoor styles; and (iii)~\textit{Sim$\to$Real}---models trained purely on simulated data are directly evaluated on real-world captures, with real-world performance feeding back to guide simulator refinement.
\textbf{Fifteen} simulated scenes and \textbf{five} real-world environments are released in this paper, providing broad scene diversity. The configurable sensor stack enables seamless retargeting to specific hardware configurations.

\begin{table}[t]
\caption{\textbf{Dataset comparison.} Comparing our \textbf{OmniOcc} Datasets with other occupancy datasets. 
Surround refers to surround-view image inputs. 
C, D, L and R denote camera and depth, LiDAR and Radar.}
\label{tab:occ3d_dataset_comparison}
\centering
\scriptsize
\setlength\tabcolsep{9pt}
\renewcommand\arraystretch{1.1}
\resizebox{\linewidth}{!}{
\begin{tabular}{lc|ccccccc}
\thickhline
\rowcolor{gray!20}
\textbf{Dataset} &Publication & \textbf{Type}  &\textbf{Surround} & \textbf{Modality} &  \textbf{\#Scenes} & \textbf{\#Frames} & \textbf{Volume Size} & \textbf{Resolution (m)} \\
\hline\hline

\rowcolor{gray!10} 
SemanticKITTI~\cite{semantickitti} &ICCV2019& Outdoor  & \ding{55} & C \& L  & 22 & 9K & [32, 256, 256]& - \\

KITTI-360~\cite{Kitti-360}&TPAMI2022 &Outdoor &\checkmark &C \& L & 11 &91K &[32, 256, 256] &[0.2, 0.2, 0.2] \\ 


\rowcolor{gray!10}
SurroundOcc~\cite{surroundocc} & ICCV2023& Outdoor & \checkmark & C \& L  & - & - &  [16, 200, 200] & [0.5, 0.5, 0.5]\\ 

OpenOccupancy~\cite{openoccupancy} &ICCV2023& Outdoor & \checkmark &  C \& L   & - & 200K &  [40, 512, 512] & -\\ 

\rowcolor{gray!10} 
Occ3D-nuScenes~\cite{occ3d} &ArXiv2023& Outdoor & \checkmark & C \& L   & 16 & 40K & [16, 200, 200] & [0.4, 0.4, 0.4]\\

Occ3D-Waymo~\cite{occ3d}& ArXiv2023& Outdoor& \checkmark & C \& L   & 14 & 200K & [128, 3200, 3200] & [0.05, 0.05, 0.05] \\

\rowcolor{gray!10}
StereoVoxelNet~\cite{li2023stereovoxelnet}&ICRA2023&Indoor\&Outdoor&\ding{55}& Stereo \& L   &-&-&-&- \\

WildOcc~\cite{WildOcc} &ArXiv2024& Outdoor & \ding{55} & C \& L   & 5  & 10K & [40, 100, 100] & [0.2, 0.2, 0.2] \\

\rowcolor{gray!10}
OmniHD-Scenes~\cite{OmniHD-Scenes}& ArXiv2024 & Outdoor & \checkmark & C \& L\& R   & 200  & 60K & - & - \\ 

Co3SOP~\cite{Co3SOP}  &ArXiv2025& Outdoor& \ding{55} & C \& L   & -  & - & [70, 1000, 1000] & [0.1, 0.1, 0.1] \\

\rowcolor{gray!10}
DSEC-SSC~\cite{DSEC-SSC} &ArXiv2025 &Outdoor & \ding{55} & S \& L   & 6 & 3K & [16, 128, 128] & [0.4, 0.4, 0.4] \\

ORAD-3D~\cite{ORAD-3D} &ArXiv2025 &Outdoor   & \checkmark & C \& L & 145  & 58K & [100, 100, 100] & - \\

\rowcolor{gray!10}
SUNCG~\cite{sscnet} &CVPR2017& Indoor  & \ding{55} & C \& D   & 46K & 140K & [144, 240, 240] & - \\ 

Occ-ScanNet~\cite{occscannet} &ECCV2024& Indoor  & \ding{55} & C \& L    & - & 65K & [36, 60, 60] & [0.08, 0.08, 0.08]\\

\rowcolor{gray!10}
EmbodiedScan~\cite{embodiedscan} &CVPR2024& Indoor  & \checkmark & C \& D  & 5,185  & 890K & [16, 40, 40] & - \\

EmbodiedOcc~\cite{embodiedocc}& ICCV2025& Indoor  & \ding{55} & C \& L    & - & 674 & [36, 60, 60] & [0.08, 0.08, 0.08]\\

\rowcolor{gray!10}
HumanoidOcc~\cite{humanoidocc}& ArXiv2025& Indoor  & \checkmark & C \& L  & - & 40K & [24, 200, 200] & [0.1, 0.1, 0.1] \\

\rowcolor{cyan!10}
\textbf{Humanoid-OmniOcc}  &\textbf{Ours}& Indoor& \checkmark& \textbf{Stereo \& D}  & 15 & 155K & [44, 384, 384] & [0.04, 0.04, 0.04]  \\

\hline\hline
\thickhline
\end{tabular}}
\end{table}

Along with the dataset, we propose the \textbf{\method} model family, which achieves state-of-the-art occupancy prediction by leveraging stereo-derived depth priors for robust 3D understanding in egocentric scenes. As shown in Tab.~\ref{tab:semantic}, our stereo-based model substantially outperforms monocular baselines on unseen test scenes while maintaining competitive results on the validation set, demonstrating stronger robustness.
Furthermore, we evaluate all methods on real-world scenes to validate sim-to-real generalization.
Comprehensive visualizations and ablation studies are provided to elucidate the properties of the proposed benchmark.

\noindent\textbf{Contributions.} Our main contributions are as follows:
\begin{itemize}
\item We present \textbf{Humanoid-OmniOcc}, the first panoramic stereo-based occupancy dataset for humanoid robots, offering full $360^\circ$ visual coverage with dense semantic labels over 15 indoor categories, spanning 15 simulated scenes and 5 real-world environments, without relying on LiDAR.

\item We develop a physically grounded data generation pipeline on Isaac~Sim following a Real2Sim2Real paradigm: sensor specifications drive simulation, simulation produces centimeter-level occupancy ground truth via multi-view stereo and geometric verification, and real-world evaluation closes the loop.

\item We propose \textbf{\method}, leveraging stereo-based depth priors to achieve state-of-the-art occupancy prediction and demonstrating strong robustness on both simulated test sets and real-world environments over monocular baselines.
\end{itemize}

\section{Related Work}

\subsection{Occupancy Perception Datasets}

Recent progress in 3D occupancy prediction has been largely driven by large-scale benchmarks~\cite{openoccupancy,surroundocc,occ3d}.
OpenOccupancy~\cite{openoccupancy} improves voxel completeness through an Augmenting and Purifying (AAP) pipeline, supported by roughly 4000 hours of human annotation and verification to ensure high-fidelity occupancy reconstruction.
SurroundOcc~\cite{surroundocc} fuses multi-frame LiDAR sweeps via Poisson reconstruction to obtain dense occupancy maps of dynamic and static scenes. 
Occ3D~\cite{occ3d} introduces two large-scale benchmarks---Occ3D-Waymo and Occ3D-nuScenes---featuring an automated annotation pipeline composed of voxel densification, occlusion reasoning, and image-guided refinement.
These efforts collectively establish strong baselines for occupancy-based 3D scene understanding, yet most remain designed for autonomous driving scenarios with meter-level resolution, limiting their direct applicability to humanoid robotics where near-field accuracy and fine-grained articulation reasoning are required.
For indoor scenes, 
EmbodiedScan~\cite{wang2024embodiedscan} introduces a multi-modal, egocentric 3D perception dataset and benchmark designed for comprehensive 3D scene understanding in embodied environments.
HumanoidOcc~\cite{humanoidocc} provides a panoramic occupancy benchmark tailored to humanoid robots, which integrates six monocular cameras and one LiDAR sensor mounted around the robot's head to achieve full-surround visual coverage with accurate spatial alignment.

\subsection{3D Occupancy Prediction}

3D occupancy prediction has emerged as a key direction in scene understanding and embodied perception.
MonoScene~\cite{cao2022monoscene} pioneered camera-based occupancy perception.
TPVFormer~\cite{huang2023tri} introduces a tri-perspective view representation framework that reconstructs 3D geometry from 2D visual cues through three orthogonal projections. 
FB-Occ~\cite{li2023fb} refines geometric reasoning through a bidirectional projection mechanism, allowing improved visibility modeling in occlusion-prone regions.
FlashOcc~\cite{yu2023flashocc} focuses on computational efficiency, compressing voxel features to achieve real-time inference without significant accuracy degradation. 
More recently, GaussianFormer~\cite{huang2024gaussianformer} and GaussianFormerV2~\cite{huang2025gaussianformer} introduce 3D semantic Gaussians to represent scene geometry and semantics.
EmbodiedOcc~\cite{wu2025embodiedocc} utilizes 3D Gaussian to represent indoor scenes in a continuous volumetric form.
Beyond pure camera-based systems, multimodal fusion frameworks have been widely explored to overcome the inherent limitations of single-sensor perception.
OccFusion~\cite{ming2024occfusion} introduces voxel-level cross-modal attention, allowing asynchronous RGB and LiDAR inputs to collaboratively predict high-density occupancy maps.
Despite notable progress, monocular-based methods often struggle with depth ambiguity and occlusion sensitivity, while LiDAR-based solutions, though accurate, are costly and difficult to deploy at scale.

\subsection{Stereo Matching}

Deep learning-based stereo matching frameworks typically \textbf{comprise} four key stages: feature extraction, cost volume construction, cost aggregation, and disparity regression.
Existing methods \textbf{are} broadly categorized into two families---accuracy-oriented and efficiency-oriented approaches.
Accuracy-focused networks~\cite{psmnet2018,gwcnet2019,leastereo,acvnet,guo2023openstereo} substantially improve disparity estimation precision by constructing high-dimensional cost volumes and performing 3D convolutional aggregation to refine matching consistency.
Meanwhile, iterative optimization-based methods~\cite{xu2023iterative,xu2024igev++,xu2023accurate,guo2023openstereo,SelectiveStereo} achieve state-of-the-art accuracy through recurrent optimization within the disparity space, further improving depth continuity and geometric fidelity.
In contrast, efficiency-oriented methods~\cite{guo2025lightstereo,bangunharcana2021coex,wang2020fadnet,deeppruner2019,shamsafar2022mobilestereonet,khamis2018stereonet} seek to reduce computational complexity by constructing compact or downsampled cost volumes, while maintaining reasonable accuracy for real-time deployment. In addition, several recent approaches have been designed specifically to improve temporal coherence in stereo video processing~\cite{jing2024matchsv,jing2024matchstereovideosbidirectional,jing2025stereovideotemporallyconsistent}, ensuring that disparities remain stable over time.
Recent foundation models such as FoundationStereo~\cite{wen2025foundationstereo}, and StereoAnything~\cite{stereoanything} further extend stereo generalization by leveraging tens of millions of diverse training pairs~\cite{stereoanything} and pretraining with depth-aware priors like DepthAnything~\cite{depthanytingv1,depthanytingv2}.
OccDepth~\cite{miao2023occdepth}, StereoScene~\cite{li2023stereoscene}, and Stereovoxelnet~\cite{li2023stereovoxelnet} incorporate stereo matching into occupancy prediction and achieve promising results. Furthermore, Cvt-Occ~\cite{ye2024cvt} introduces a temporal cost volume framework tailored for multi-view and multi-frame occupancy prediction.
However, these methods are limited to front-view stereo settings, and no surround-stereo occupancy prediction framework has been explored due to the lack of corresponding datasets.

\section{Humanoid-OmniOcc Dataset}

\subsection{Real2Sim2Real Design Philosophy}
\label{subsec:real2sim2real}

Humanoid-OmniOcc is built around a \textbf{Real2Sim2Real} closed-loop paradigm that tightly couples simulation with real-world deployment, as illustrated below.

\noindent\textbf{Real$\boldsymbol{\to}$Sim.}
The physical sensor configuration of the Unitree~G1 robot---including stereo camera intrinsics, extrinsics, baseline ($6$\,cm), field of view, and mounting geometry---is precisely replicated in NVIDIA Isaac Sim. Photorealistic physically-based rendering (PBR) with calibrated material properties and lighting further minimizes the visual domain gap, ensuring that the simulated observations closely match real sensor outputs.

\noindent\textbf{Sim (Data Generation).}
With the sensor-accurate digital twin in place, we generate large-scale training data with pixel-accurate dense annotations that would be prohibitively expensive to obtain in the real world. The simulator produces centimeter-level occupancy ground truth, per-voxel semantic labels across 15 categories, and metrically calibrated depth maps---all automatically and at scale (155K+ samples across 15 diverse indoor scenes).

\noindent\textbf{Sim$\boldsymbol{\to}$Real.}
Models trained exclusively on simulated data are deployed and evaluated on real-world captures from the same Unitree~G1 platform in five distinct indoor environments (Bar, Corridor, Office, Apartment). The real-world evaluation quantitatively measures the sim-to-real transfer gap and reveals domain-specific failure modes (\eg, lighting discrepancies, material reflectance mismatch), which in turn inform iterative refinement of the simulation parameters---closing the Real2Sim2Real loop.

This closed-loop design ensures that the benchmark is not a one-off simulation artifact but a continuously improvable pipeline tightly coupled with real-world deployment requirements.

\subsection{Virtual Scene Setup}

We construct diverse virtual indoor environments using \textit{Isaac Sim}
to emulate realistic object layouts, illumination conditions, and physical interactions,
providing a controllable and reproducible foundation for data collection.
The scenes span a wide range of global interior styles---including Scandinavian minimalism,
Baroque luxury, and modern industrial design---capturing the diversity of real-world indoor environments
across different cultural and aesthetic contexts.
Each environment contains static objects (\eg, tables, chairs, cabinets, walls),
all assigned physically accurate attributes including material reflectivity, collision dynamics,
and lighting parameters, ensuring faithful simulation of real-world photometric and geometric behavior.

\paragraph{Data quality, diversity, and scalability.}
We prioritize sim-to-real fidelity via photorealistic physically-based rendering (PBR) and sensor-accurate camera emulation (intrinsics/extrinsics, noise, FoV, baseline), enabling direct retargeting to specific hardware.
Ground-truth occupancy is produced at centimeter-level precision via multi-view geometric verification, yielding high-accuracy labels at low cost with high collection throughput.

\begin{wrapfigure}{r}{0.5\linewidth}
    \centering
    \includegraphics[width=0.88\linewidth]{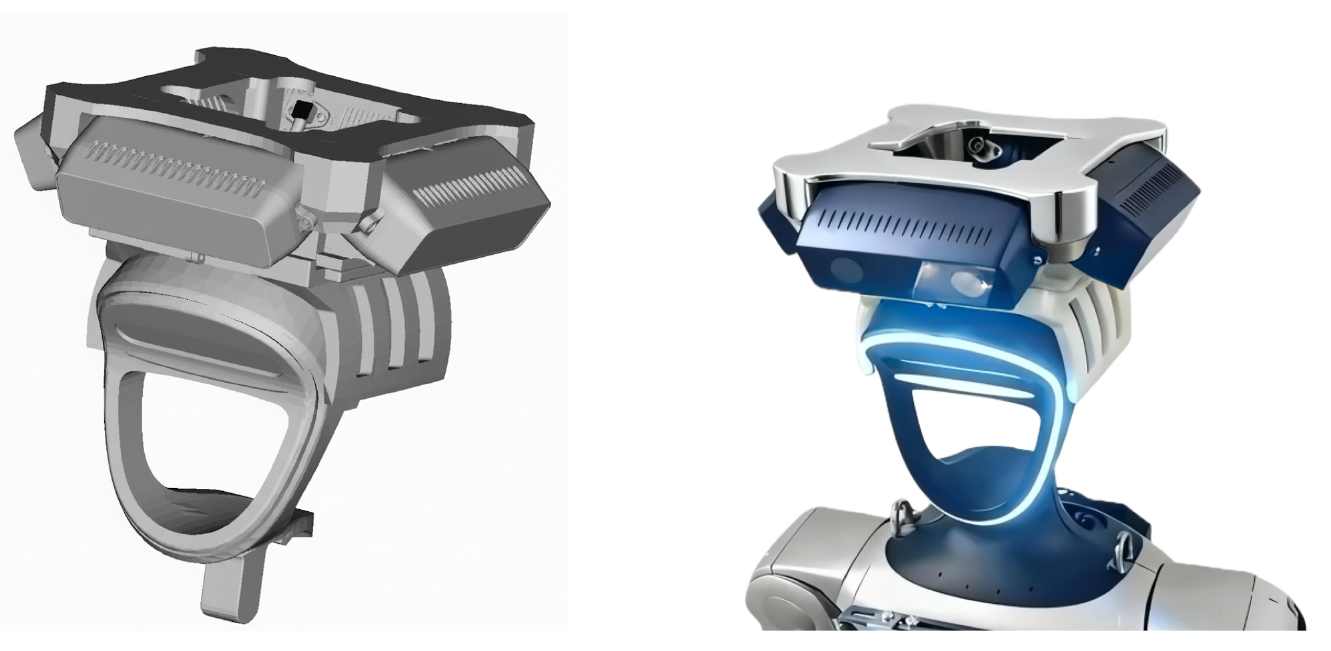}
    \caption{The data collection setup, featuring the CAD model of the stereo head (left) alongside its integration into the physical humanoid robot (right).}
    \label{fig:sensor}
\end{wrapfigure}

\subsection{Omnidirectional Stereo Configuration}

As shown in Fig.~\ref{fig:sensor}, four stereo camera modules are mounted on the Unitree G1's head facing the \textit{front}, \textit{rear}, \textit{left}, and \textit{right} directions, forming a panoramic stereo perception system. Each stereo pair performs depth estimation through disparity computation, ensuring $360^\circ$ spatial coverage. 
In our setup, we employ stereo cameras with a resolution of $1280\times1080$, a baseline of $6$ cm, and a focal length of $596.81$ pixels.
The field of view (FOV) of the raw distorted images is approximately $106^{\circ}$ horizontally and $86^{\circ}$ vertically ($\pm3^{\circ}$),
which becomes $93^{\circ}$ (H) and $83^{\circ}$ (V) ($\pm3^{\circ}$) after stereo rectification.
Intrinsic and extrinsic parameters of all cameras are calibrated and recorded to support precise spatial alignment.



\subsection{Autonomous Data Collection via Dynamic Window Approach}
To capture realistic motion and viewpoint diversity, an autonomous navigation policy based on the Dynamic Window Approach (DWA) is integrated into the robot control system. DWA samples feasible linear and angular velocity commands under velocity and acceleration constraints, predicts short-term trajectories, and scores them according to goal alignment, obstacle clearance, and motion smoothness. The optimal velocity command is executed to ensure safe, collision-free navigation. During navigation, synchronized stereo image pairs and robot poses are continuously recorded for subsequent ground-truth generation. 

An overview of the Humanoid-OmniOcc dataset, ground truth generation in both simulation and the real world, as well as additional analyses of the dataset, are provided in the \textcolor{red}{\textit{Supplementary Material}}.

\section{\method Model}

\subsection{Task Definition}

\begin{figure}[!t]
    \centering
    \includegraphics[width=\linewidth]{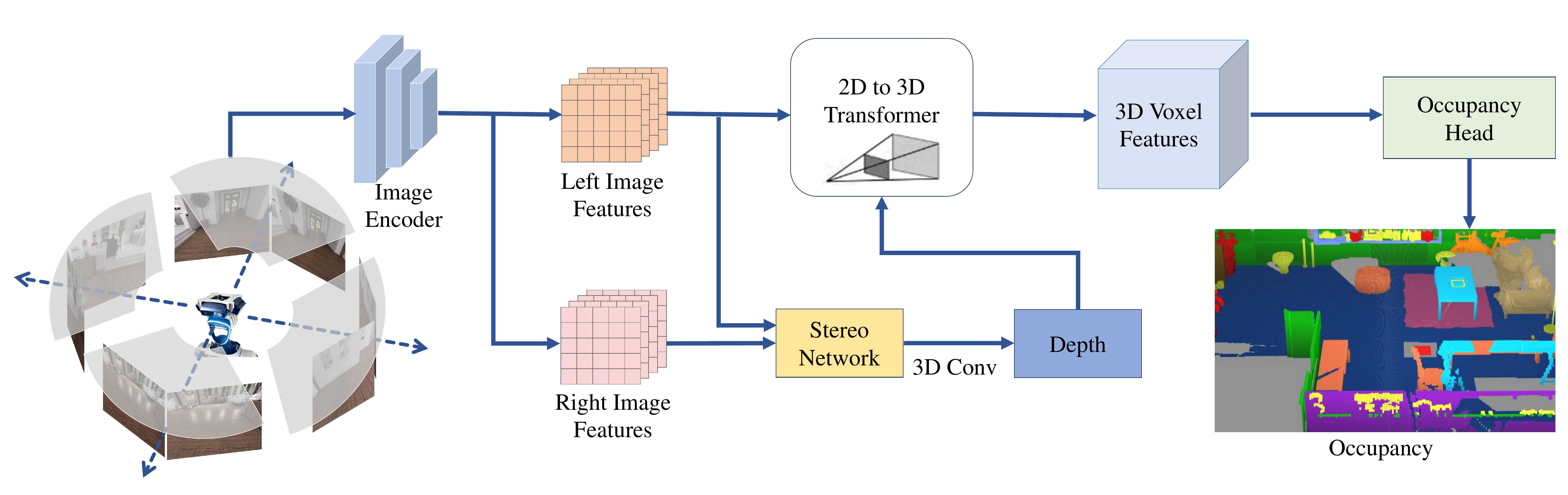}
    \caption{\textbf{The pipeline of our proposed \method~framework.}}
    \label{fig:algorithm}
\end{figure}

As shown in Figure~\ref{fig:algorithm}, given synchronized multi-view stereo inputs, the objective of our task is to infer the 3D occupancy state of each voxel within the surrounding environment. 
Specifically, the model receives frames captured from four stereo pairs positioned at the front, back, left, and right of the humanoid platform, denoted as 
$\{I_{i}^{l}, I_{i}^{r}\} \subset \mathbb{R}^{H_i \times W_i \times 3}$, where $i = 1,...,4$. 
Each stereo pair is calibrated with known intrinsics $\{\mathbf{K}_i^{l}, \mathbf{K}_i^{r}\}$ and extrinsics $\{\mathbf{T}_i^{l}, \mathbf{T}_i^{r}\}$ relative to the robot frame. 
The network predicts a volumetric occupancy grid $\hat{O} \in \{0,1\}^{X \times Y \times Z}$, 
where each voxel represents an \textit{occupancy state} (free, occupied).

\subsection{Feature Extraction}
\label{subsec:feature}

\paragraph{Multi-view encoding.}
Eight synchronized RGB images from four stereo pairs (\textit{front}, \textit{back}, \textit{left}, and \textit{right}) are encoded by a shared 2D backbone $\mathcal{E}$ into multi-scale feature maps $\{F_i^l, F_i^r\}$, 
where $i \in \{1,2,3,4\}$ corresponds to the four viewing directions, 
and the superscripts $l$ and $r$ denote the left and right images of each stereo pair, respectively.

\subsection{Stereo Depth Network}
\label{subsec:stereo}

To model pixel correspondences along potential disparities, a differentiable cost volume $C_i^{\text{disp}}$ is constructed over disparity hypotheses $d \in [0, D_{\max}]$ as:
\begin{equation}
C_i^{\text{disp}}(u,v,d) = \big\lVert F_i^l(u,v) - F_i^r(u,v+d) \big\rVert_{1},
\end{equation}
where $C_i^{\text{disp}}(u,v,d,:)$ encodes the similarity between left and right feature embeddings.

Disparity-based cost volumes suffer from non-uniform depth sensitivity, as the same disparity shift corresponds to larger depth variations at greater distances.
To alleviate this issue, we transform $C_i^{\text{disp}}$ into a depth-oriented cost representation $C_i^{\text{depth}}$ using bilinear interpolation guided by the depth-to-disparity relation.
This reformulation aligns the convolutional operations with actual geometric depth, ensuring equal treatment across different distance ranges.

The depth cost volume $C_i^{\text{depth}}$ is processed by a series of 3D convolutions 
to generate the aggregated depth feature volume $S_i^{\text{depth}}$.
$S_i^{\text{depth}}$ is then used to regress pixel-wise depth via a softmax-weighted expectation:
\begin{equation}
P_i(u,v,z) = \sigma\!\big(-S_i^{\text{depth}}(u,v,z)\big),
\label{eq:depth_prob}
\end{equation}
\begin{equation}
Z_i^l(u,v) = \sum_{z=1}^{K} P_i(u,v,z)\, z,
\label{eq:depth_prediction}
\end{equation}
where $Z_i^l(u,v)$ denotes the predicted metric depth of the left image.
This depth-based formulation leads to smoother and more stable predictions, particularly for far-range regions.

\subsection{2D-to-3D Transformer}
\label{subsec:lift}

We lift only \emph{left-view} features into a robot-centric voxel grid $\mathcal{V}$ using depth-discretized splatting guided by the left camera's stereo-derived depth posterior.

Let $\mathcal{I}=\{1,2,3,4\}$ index the four stereo rigs (front/back/left/right). For the left camera of rig $i$, denote its intrinsics and extrinsics by $\mathbf{K}_i^{l}$ and $\mathbf{T}_i^{l}$ (robot frame), and the left-view feature map by $F_i^{l}$. For a pixel $\mathbf{x}=(u,v)$ and discretized depth bins $\{z_k\}_{k=1}^{K}$, the back-projected 3D point in the robot frame is
\begin{equation}
\mathbf{X}_w^{\,l}(\mathbf{x},z_k)=\mathbf{T}_i^{l}\,\Pi^{-1}\!\big(\mathbf{x},z_k,\mathbf{K}_i^{l}\big).
\end{equation}
With trilinear splatting $\phi(\cdot\!\rightarrow\!v)$ and a feature projector $\mathcal{P}$, the camera-to-voxel aggregation becomes
\begin{equation}
\label{eq:forward_lift_left}
V_{\text{cam}}(v)=\sum_{i\in\mathcal{I}}\;\sum_{\mathbf{x}}\;\sum_{k=1}^{K}
q_i(\mathbf{x},k)\,\phi\!\Big(\mathbf{X}_w^{\,l}(\mathbf{x},z_k)\!\rightarrow\! v\Big)\,\mathcal{P}\!\big(F_i^{l}(\mathbf{x})\big),
\end{equation}
where $q_i(\mathbf{x},k)$ is the \emph{left-view} stereo depth posterior from Sec.~\ref{subsec:stereo}. A lightweight transformer refines $V_{\text{cam}}$ by aggregating multi-view context along camera rays.

\subsection{Occupancy Head and Loss}
\label{subsec:head_train}

The occupancy head consumes the lifted voxel feature map $V_{\text{cam}} \in \mathbb{R}^{X \times Y \times Z \times D}$ and predicts per-voxel logits $\mathbf{o}(v) \in \mathbb{R}^{C_{\text{occ}}}$ using a lightweight 3D decoder composed of sparse convolution and trilinear upsampling layers. A softmax classifier converts logits to occupancy probabilities.
Following FB-Occ~\cite{li2023fb}, we supervise the predicted occupancy map $\hat{O}$ with a weighted composite loss:
\begin{equation}
\mathcal{L}_{\text{occ}} =
\mathcal{L}_{\text{focal}} +
\mathcal{L}^{\text{geo}}_{\text{scal}} +
 \mathcal{L}^{\text{sem}}_{\text{scal}} +
 \mathcal{L}_{\text{lovasz}}.
\end{equation}

In parallel, the stereo depth prediction (Sec.~\ref{subsec:stereo}) is supervised with binary cross-entropy over discretized depth bins. 
The full objective function couples voxel occupancy and stereo depth supervision:

\begin{equation}
\mathcal{L}_{\text{total}} =
\mathcal{L}_{\text{occ}} +
\lambda \cdot \mathcal{L}_{\text{depth}}.
\end{equation}

This joint optimization framework aligns appearance, depth, and volumetric reasoning. Notably, the stereo posterior 
$q_i(\mathbf{x},k)$ (Eq.~\ref{eq:forward_lift_left})
guides voxel lifting, while occupancy gradients implicitly regularize depth prediction by enforcing geometric consistency within 3D space.

\section{Experiments}

\subsection{Implementation Details.}

All experiments are conducted on a cluster equipped with 16 NVIDIA H20 GPUs.
The network is trained for 10 epochs using the AdamW optimizer with a learning rate of $4\times10^{-4}$.
We set the batch size to $32$.
All input images are resized to $544\times640$ before training and inference.
For LightStereo~\cite{guo2025lightstereo}, COEX~\cite{bangunharcana2021coex}, and IGEV~\cite{xu2023iterative} stereo models, we initialize the models with StereoAnything~\cite{stereoanything} pretrained weights to leverage large-scale stereo correspondence priors.
For FoundationStereo~\cite{wen2025foundationstereo}, we adopt the official pretrained weights released by the authors. 
We adapt FB-Occ~\cite{li2023fb}, FlashOcc~\cite{yu2023flashocc}, and SurroundOcc~\cite{surroundocc} (originally designed for nuScenes with 6 surround-view cameras) to a 4-view input setting using the left images from each stereo pair.

\subsection{Evaluation Metrics}

Following SurroundOcc~\cite{surroundocc}, we evaluate semantic occupancy using the per-class intersection-over-union (IoU) and the mean IoU over valid classes (mIoU), which jointly measure geometric consistency and fine-grained semantic recognition. 

\begin{table}[t]
\centering
\caption{Semantic occupancy prediction results on the  
\textbf{test} sets and in the real world. We report per-class IoU and mIoU (\%).} 
\label{tab:semantic}
\resizebox{0.95\textwidth}{!}{%
\begin{tabular}{l|c|cc|ccccccccccccc}
\toprule
Method &Input & IoU & mIoU
& \rotatebox{90}{\textcolor{floor}{$\blacksquare$} floor}
& \rotatebox{90}{\textcolor{partitions}{$\blacksquare$} partitions}
& \rotatebox{90}{\textcolor{door}{$\blacksquare$} door}
& \rotatebox{90}{\textcolor{chair}{$\blacksquare$} chair}
& \rotatebox{90}{\textcolor{table}{$\blacksquare$} table}
& \rotatebox{90}{\textcolor{sofa}{$\blacksquare$} sofa}
& \rotatebox{90}{\textcolor{bed}{$\blacksquare$} bed}
& \rotatebox{90}{\textcolor{appliance}{$\blacksquare$} appliance}
& \rotatebox{90}{\textcolor{cabinet}{$\blacksquare$} cabinet}
& \rotatebox{90}{\textcolor{carpet}{$\blacksquare$} carpet}
& \rotatebox{90}{\textcolor{plant}{$\blacksquare$} plant}
& \rotatebox{90}{\textcolor{objects}{$\blacksquare$} objects}
& \rotatebox{90}{\textcolor{suspended_object}{$\blacksquare$} suspended} \\
\midrule
\multicolumn{17}{c}{\cellcolor{gray!10}\textit{Test Set}} \\
\midrule
FB-Occ~\cite{li2023fb}        & Mono   & 28.59	& 5.11	& 36.76	& 6.07	& 4.44	& 2.24	& 2.32	& 0.82	& 1.43	& 0.00	& 4.27	& 2.31	& 4.13	& 1.68	& 0.01 \\
FlashOcc~\cite{yu2023flashocc}      & Mono  & 18.05	& 1.71	& 16.69	& 1.87	& 1.72	& 0.08	& 0.09	& 0.05	& 0.00	& 0.02	& 0.56	& 0.06	& 0.01	& 1.14	& 0.00 \\
SurroundOcc~\cite{surroundocc}   & Mono   & 24.71	& 6.86	& 28.22	& 8.10	& 2.17	& 4.62	& 14.65	& 7.96	& 5.93	& 0.01	& 7.81	& 2.34	& 2.88	& 4.53	& 0.01 \\
GaussianFormer~\cite{huang2024gaussianformer}   & Mono   & 26.15	& 5.84	& 33.43	& 5.03	& 1.20	& 4.47	& 6.54	& 2.38	& 5.19	& 0.09	& 6.11	& 5.31	& 3.09	& 3.11	& 0.00 \\
\rowcolor{blue!5}
HS$^2$Occ (Ours) & Stereo & 29.67	& 11.69	& 50.85	& 4.50	& 12.97	& 5.17	& 6.95	& 9.57	& 9.77	& 7.05	& 16.71	& 8.02	& 10.58	& 9.70	& 0.12 \\
\midrule
\multicolumn{17}{c}{\cellcolor{gray!10}\textit{Real World}} \\
\midrule
FB-Occ~\cite{li2023fb}         & Mono   & 12.22	& 2.42	& 8.30	& 3.92	& 0.00	& 3.90	& 4.54	& 4.88	& 0.00	& 0.00	& 2.41	& 1.61	& 0.23	& 1.59	& 0.04 \\
FlashOcc~\cite{yu2023flashocc}      & Mono   & 15.39	& 2.42	& 18.21	& 4.58	& 0.00	& 1.58	& 0.82	& 1.63	& 0.00	& 0.00	& 0.91	& 0.26	& 1.29	& 0.88	& 1.36 \\
SurroundOc~\cite{surroundocc}   & Mono   & 20.35	& 8.89	& 26.76	& 10.18	& 1.48	& 10.72	& 10.63	& 16.21	& 18.23	& 0.06	& 4.81	& 0.09	& 1.44	& 4.45	& 10.47 \\
GaussianFormer~\cite{huang2024gaussianformer}   & Mono   & 17.11	& 5.30	& 21.55	& 7.22	& 0.86	& 9.32	& 6.98	& 6.51	& 1.90	& 0.00	& 2.75	& 1.40	& 2.14	& 6.25	& 2.01 \\
\rowcolor{blue!5}
HS$^2$Occ (Ours) & Stereo & 35.45	& 19.26	& 51.31	& 14.29	& 7.77	& 22.31	& 22.43	& 34.74	& 34.93	& 1.64	& 13.52	& 11.40	& 11.57	& 12.12	& 12.34 \\
\bottomrule
\end{tabular}%
}
\end{table}

\subsection{Main Results}
\label{sec:main_results}

Table~\ref{tab:semantic} summarizes voxel-level semantic occupancy prediction on the Humanoid-OmniOcc \textbf{test} set and real-world captures. We compare our proposed \textbf{\method} with four SOTA monocular occupancy prediction models: FB-Occ~\cite{li2023fb} (CVPR'23 3D occupancy prediction champion method), FlashOcc~\cite{yu2023flashocc}, SurroundOcc~\cite{surroundocc} and GaussianFormer~\cite{huang2024gaussianformer}.
On the test set, our stereo-based \textbf{HS$^2$Occ} achieves the best overall performance with 29.67 IoU and 11.69 mIoU, outperforming the strongest monocular baseline (FB-Occ: 28.59 IoU / 5.11 mIoU), indicating notably better semantic completeness on indoor long-tail categories. More importantly, HS$^2$Occ generalizes strongly to real-world scenes under the Real2Sim2Real protocol, reaching 35.45 IoU and 19.26 mIoU, which exceeds the best monocular results (best IoU: 15.39, best mIoU: 5.34). 
Per-class results further show that stereo depth priors substantially improve interaction-relevant semantics and thin structures: on the test set, HS$^2$Occ boosts \textit{floor} (50.85), \textit{door} (12.97), and \textit{cabinet} (16.71), while maintaining non-trivial IoUs for object-centric classes where monocular methods often collapse; in real-world evaluation, HS$^2$Occ achieves strong recognition for \textit{chair/table/sofa/bed} (22.31/22.43/34.74/34.93), validating robust 2D-to-3D lifting under domain shift. 

\subsection{Ablation study}

Our dataset and model primarily focus on the stereo-based setting, aiming to explore how binocular perception contributes to accurate and stable occupancy prediction. Therefore, we keep the occupancy head and training protocol fixed, and ablate the stereo component by plugging in different stereo backbones while using the same surround stereo inputs and 2D-to-3D lifting pipeline. 

\textbf{Ablation study on Stereo module.} As shown in Tab.~\ref{tab:ablationStereo},
FoundationStereo~\cite{wen2025foundationstereo} achieves the best overall semantic occupancy performance, particularly on real-world data, demonstrating that stronger and more robust disparity estimation can substantially improve downstream occupancy prediction quality.
Overall, these results confirm that occupancy quality is highly sensitive to stereo depth reliability: a stronger stereo model can significantly enhance semantic completeness and real-world generalization, motivating our use of FoundationStereo~\cite{wen2025foundationstereo} as the default stereo backbone.

\textbf{Ablation Study on Disparity-to-Depth Formulation.}
As shown in Tab.~\ref{tab:ablationStereoType}, compared with DDVM, SDN consistently improves occupancy on both the test set (IoU/mIoU: 28.04/10.07 $\rightarrow$ 29.67/11.69) and real-world data (32.80/17.03 $\rightarrow$ 35.45/19.26). 

\begin{table}[t]
  \small
  \centering
  \caption{Ablation Study on Stereo module.}
  \label{tab:ablationStereo}
  \resizebox{0.95\textwidth}{!}{%
  \begin{tabular}{l|cc|ccccccccccccc}
  \toprule
  Stereo &IoU& mIoU 
  & \rotatebox{90}{\textcolor{floor}{$\blacksquare$} floor}
& \rotatebox{90}{\textcolor{partitions}{$\blacksquare$} partitions}
& \rotatebox{90}{\textcolor{door}{$\blacksquare$} door}
& \rotatebox{90}{\textcolor{chair}{$\blacksquare$} chair}
& \rotatebox{90}{\textcolor{table}{$\blacksquare$} table}
& \rotatebox{90}{\textcolor{sofa}{$\blacksquare$} sofa}
& \rotatebox{90}{\textcolor{bed}{$\blacksquare$} bed}
& \rotatebox{90}{\textcolor{appliance}{$\blacksquare$} appliance}
& \rotatebox{90}{\textcolor{cabinet}{$\blacksquare$} cabinet}
& \rotatebox{90}{\textcolor{carpet}{$\blacksquare$} carpet}
& \rotatebox{90}{\textcolor{plant}{$\blacksquare$} plant}
& \rotatebox{90}{\textcolor{objects}{$\blacksquare$} objects}
& \rotatebox{90}{\textcolor{suspended_object}{$\blacksquare$} suspended} \\
  \midrule
  \multicolumn{16}{c}{\cellcolor{gray!10}\textit{Test Set}} \\
  \midrule
  LStereo-S~\cite{guo2025lightstereo} & 33.46	& 6.55	& 39.33	& 6.06	& 0.00	& 6.31	& 12.37	& 3.40	& 6.44	& 0.21	& 5.37	& 0.14	& 1.64	& 3.87	& 0.00 \\
  LStereo-L~\cite{guo2025lightstereo} & 35.11	& 8.09	& 42.73	& 6.62	& 0.08	& 7.26	& 15.21	& 4.49	& 9.25	& 0.03	& 6.51	& 0.88	& 8.54	& 3.57	& 0.00 \\
  COEX~\cite{bangunharcana2021coex} & 31.17	& 5.20	& 43.24	& 2.49	& 0.01	& 5.14	& 7.40	& 0.54	& 1.88	& 0.02	& 3.46	& 0.31	& 0.57	& 2.51	& 0.00 \\
  IGEV~\cite{xu2023iterative} & 27.59	& 5.59	& 36.01	& 2.98	& 0.02	& 6.56	& 10.88	& 2.52	& 0.47	& 0.07	& 2.97	& 0.09	& 6.58	& 3.57	& 0.00 \\
  \rowcolor{blue!5}
  FDS~\cite{wen2025foundationstereo} & 29.67	& 11.69	& 50.85	& 4.50	& 12.97	& 5.17	& 6.95	& 9.57	& 9.77	& 7.05	& 16.71	& 8.02	& 10.58	& 9.70	& 0.12 \\
  \midrule
  \multicolumn{16}{c}{\cellcolor{gray!10}\textit{Real World}} \\
  \midrule
  LStereo-S~\cite{guo2025lightstereo} & 33.69	& 9.00	& 44.86	& 13.61	& 3.81	& 5.33	& 14.88	& 16.08	& 0.86	& 0.17	& 5.43	& 2.62	& 0.99	& 7.54	& 0.76 \\
  LStereo-L~\cite{guo2025lightstereo} & 33.66	& 9.78	& 43.52	& 18.51	& 0.18	& 11.32	& 13.88	& 15.73	& 2.87	& 0.05	& 6.00	& 0.33	& 7.01	& 7.37	& 0.36 \\
  COEX~\cite{bangunharcana2021coex} & 30.07	& 6.96	& 43.48	& 11.78	& 0.06	& 6.30	& 12.94	& 5.90	& 0.44	& 0.04	& 1.82	& 0.00	& 1.65	& 5.95	& 0.07 \\
  IGEV~\cite{xu2023iterative} & 28.22	& 9.32	& 42.41	& 9.55	& 0.65	& 10.70	& 13.20	& 27.65	& 0.00	& 0.98	& 4.34	& 0.19	& 4.66	& 6.89	& 0.00 \\
  \rowcolor{blue!5}
  FDS~\cite{wen2025foundationstereo} & 35.45	& 19.26	& 51.31	& 14.29	& 7.77	& 22.31	& 22.43	& 34.74	& 34.93	& 1.64	& 13.52	& 11.40	& 11.57	& 12.12	& 12.34 \\
  \bottomrule
  \end{tabular}%
  }
\end{table}

\begin{table}[t]
  \small
  \centering
  \caption{Ablation Study on Disparity-to-Depth: DDVM~\cite{chen2025unleashing} vs. SDN.}
  \label{tab:ablationStereoType}
  \resizebox{0.95\textwidth}{!}{%
  \begin{tabular}{l|c|cc|ccccccccccccc}
  \toprule
  Stereo&Disp2Depth &IoU& mIoU 
  & \rotatebox{90}{\textcolor{floor}{$\blacksquare$} floor}
& \rotatebox{90}{\textcolor{partitions}{$\blacksquare$} partitions}
& \rotatebox{90}{\textcolor{door}{$\blacksquare$} door}
& \rotatebox{90}{\textcolor{chair}{$\blacksquare$} chair}
& \rotatebox{90}{\textcolor{table}{$\blacksquare$} table}
& \rotatebox{90}{\textcolor{sofa}{$\blacksquare$} sofa}
& \rotatebox{90}{\textcolor{bed}{$\blacksquare$} bed}
& \rotatebox{90}{\textcolor{appliance}{$\blacksquare$} appliance}
& \rotatebox{90}{\textcolor{cabinet}{$\blacksquare$} cabinet}
& \rotatebox{90}{\textcolor{carpet}{$\blacksquare$} carpet}
& \rotatebox{90}{\textcolor{plant}{$\blacksquare$} plant}
& \rotatebox{90}{\textcolor{objects}{$\blacksquare$} objects}
& \rotatebox{90}{\textcolor{suspended_object}{$\blacksquare$} suspended} \\
  \midrule
  \multicolumn{17}{c}{\cellcolor{gray!10}\textit{Test Set}} \\
  \midrule
  FDS~\cite{wen2025foundationstereo} & DDVM & 28.04	& 10.07	& 50.87	& 6.55	& 4.85	& 6.39	& 13.95	& 7.31	& 3.29	& 0.18	& 6.65	& 17.74	& 6.36	& 6.80	& 0.00 \\
  \rowcolor{gray!10}
  FDS~\cite{wen2025foundationstereo} & SDN & 29.67	& 11.69	& 50.85	& 4.50	& 12.97	& 5.17	& 6.95	& 9.57	& 9.77	& 7.05	& 16.71	& 8.02	& 10.58	& 9.70	& 0.12 \\
  \midrule
  \multicolumn{17}{c}{\cellcolor{gray!10}\textit{Real World}} \\
  \midrule
  FDS~\cite{wen2025foundationstereo} & DDVM & 32.80	& 17.03	& 47.92	& 18.06	& 5.95	& 24.84	& 22.29	& 29.58	& 31.15	& 0.82	& 5.56	& 4.21	& 14.53	& 10.99	& 5.45 \\
  \rowcolor{gray!10}
  FDS~\cite{wen2025foundationstereo} & SDN & 35.45	& 19.26	& 51.31	& 14.29	& 7.77	& 22.31	& 22.43	& 34.74	& 34.93	& 1.64	& 13.52	& 11.40	& 11.57	& 12.12	& 12.34 \\
  \bottomrule
  \end{tabular}%
  }
\end{table}

\begin{figure}[!t]
    \centering
    \begin{adjustbox}{rotate=90}\hspace{3mm}Testset\end{adjustbox}
    \begin{subfigure}[b]{0.15\textwidth}
        \centering
        \begin{minipage}[b][0.8\textwidth][t]{0.8\textwidth}
            \includegraphics[width=0.5\textwidth,height=0.5\textwidth]{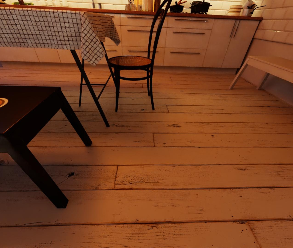}%
            \includegraphics[width=0.5\textwidth,height=0.5\textwidth]{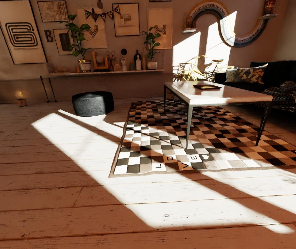}\\[-0.5mm]
            \includegraphics[width=0.5\textwidth,height=0.5\textwidth]{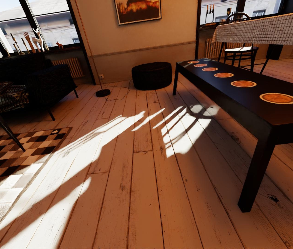}%
            \includegraphics[width=0.5\textwidth,height=0.5\textwidth]{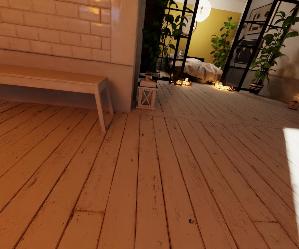}
        \end{minipage}
    \end{subfigure}\hspace{-2mm}
    \begin{subfigure}[b]{0.15\textwidth}
        \centering
        \includegraphics[width=\textwidth, height=0.8\textwidth]{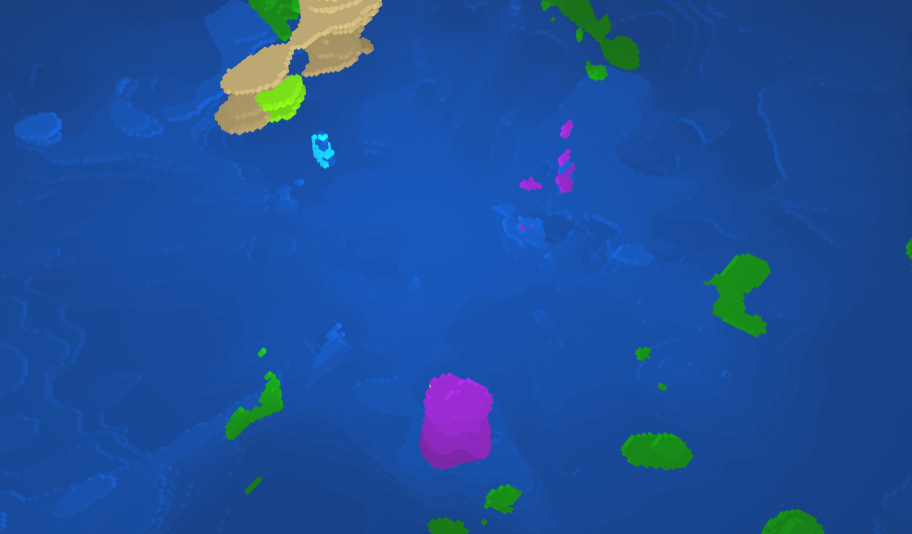}
    \end{subfigure}
    \begin{subfigure}[b]{0.15\textwidth}
        \centering
        \includegraphics[width=\textwidth, height=0.8\textwidth]{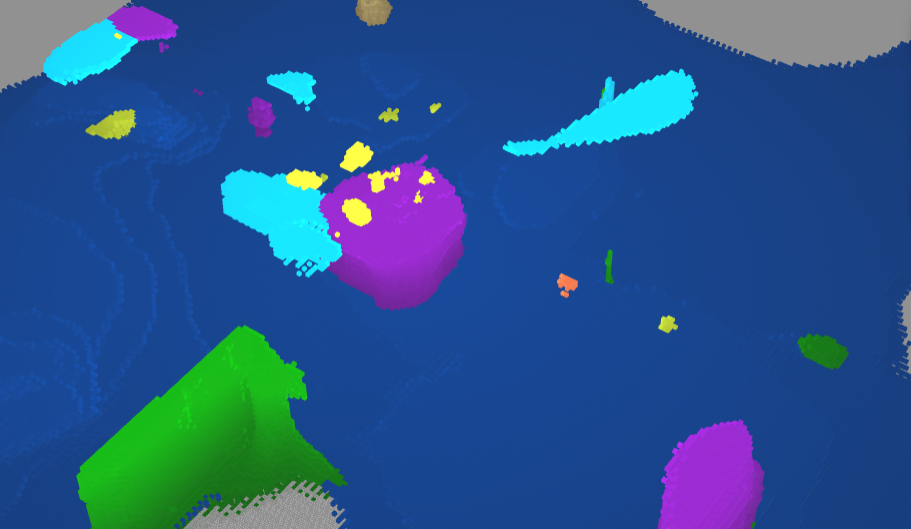}
    \end{subfigure}
    \begin{subfigure}[b]{0.15\textwidth}
        \centering
        \includegraphics[width=\textwidth, height=0.8\textwidth]{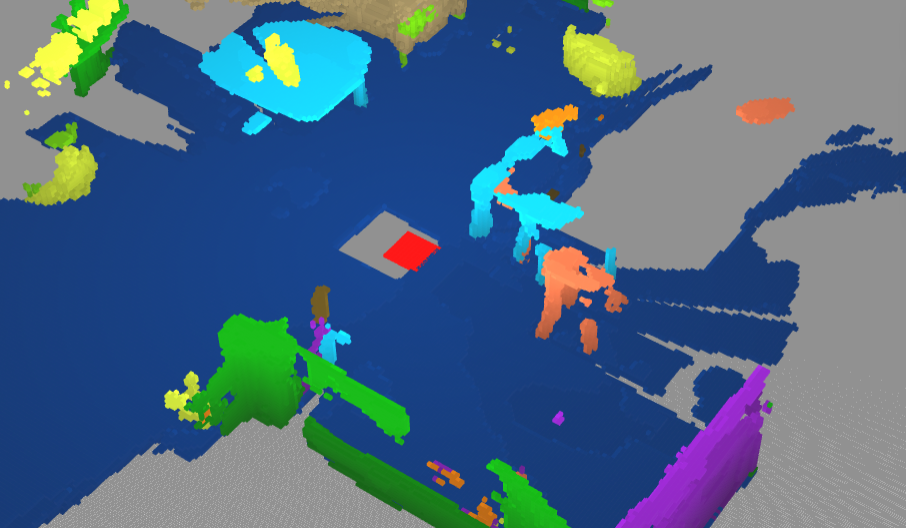}
    \end{subfigure}
    \begin{subfigure}[b]{0.15\textwidth}
        \centering
        \includegraphics[width=\textwidth, height=0.8\textwidth]{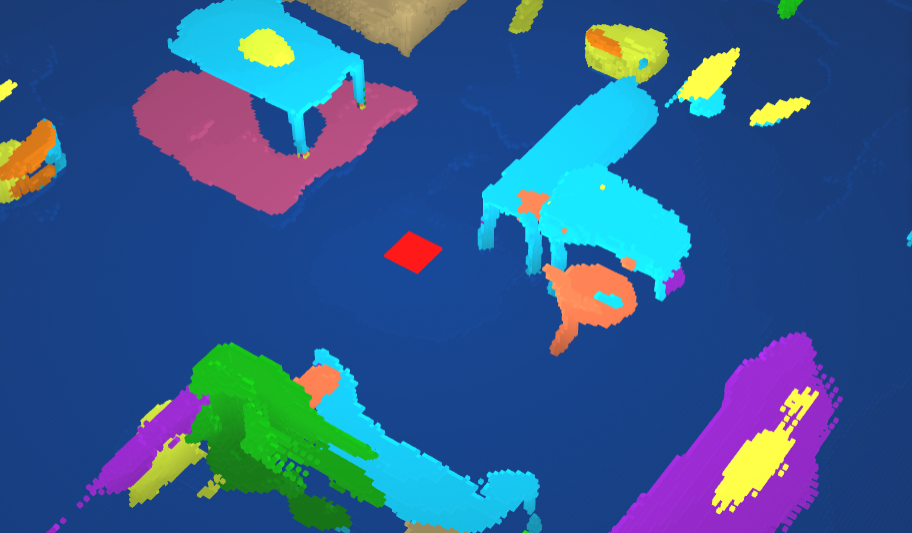}
    \end{subfigure}
    \begin{subfigure}[b]{0.15\textwidth}
        \centering
        \includegraphics[width=\textwidth, height=0.8\textwidth]{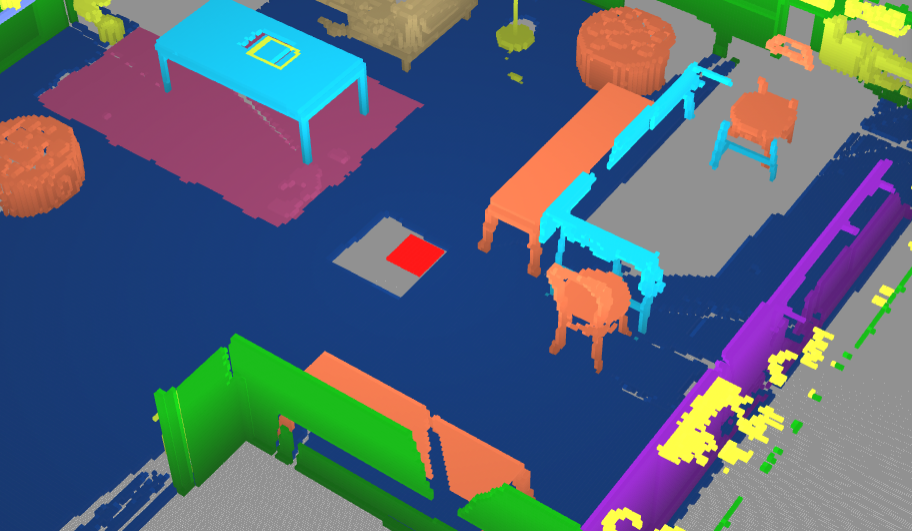}
    \end{subfigure}

    \vspace{1mm}
    \begin{adjustbox}{rotate=90}\hspace{3mm}Office\end{adjustbox}
    \begin{subfigure}[b]{0.15\textwidth}
        \centering
        \begin{minipage}[b][0.8\textwidth][t]{0.8\textwidth}
            \includegraphics[width=0.5\textwidth,height=0.5\textwidth]{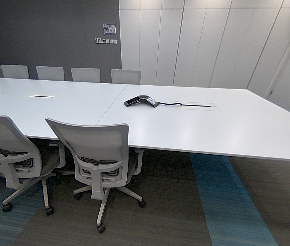}%
            \includegraphics[width=0.5\textwidth,height=0.5\textwidth]{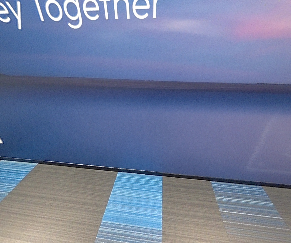}\\[-0.5mm]
            \includegraphics[width=0.5\textwidth,height=0.5\textwidth]{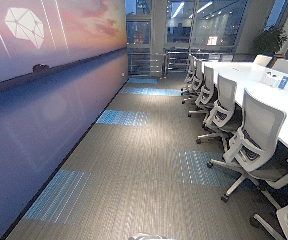}%
            \includegraphics[width=0.5\textwidth,height=0.5\textwidth]{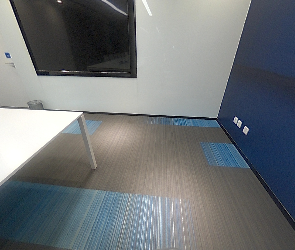}
        \end{minipage}
    \end{subfigure}\hspace{-2mm}
    \begin{subfigure}[b]{0.15\textwidth}
        \centering
        \includegraphics[width=\textwidth, height=0.8\textwidth]{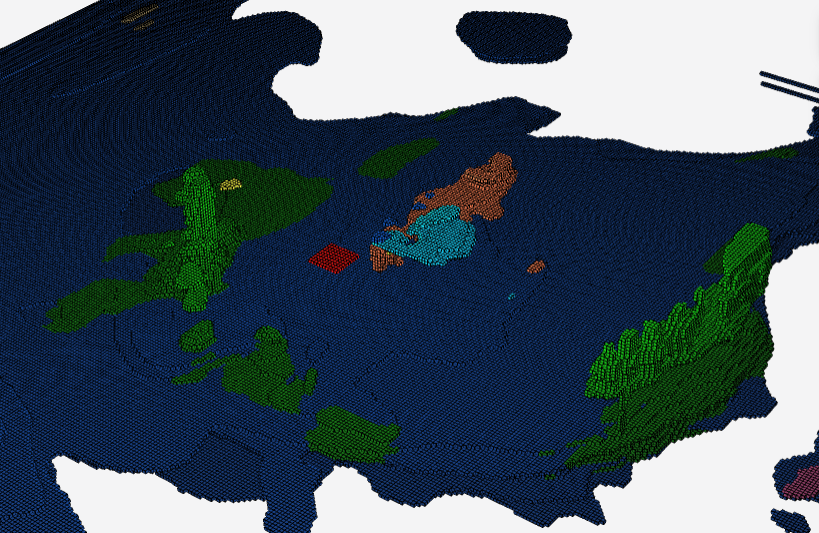}
    \end{subfigure}
    \begin{subfigure}[b]{0.15\textwidth}
        \centering
        \includegraphics[width=\textwidth, height=0.8\textwidth]{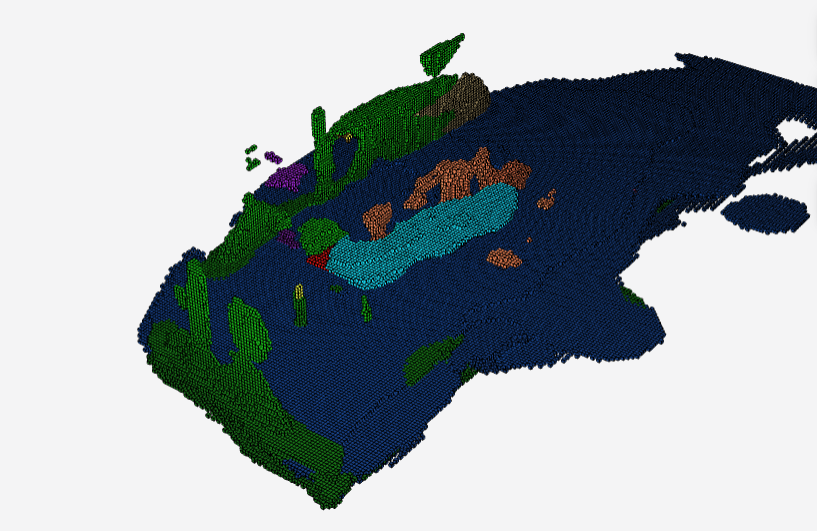}
    \end{subfigure}
    \begin{subfigure}[b]{0.15\textwidth}
        \centering
        \includegraphics[width=\textwidth, height=0.8\textwidth]{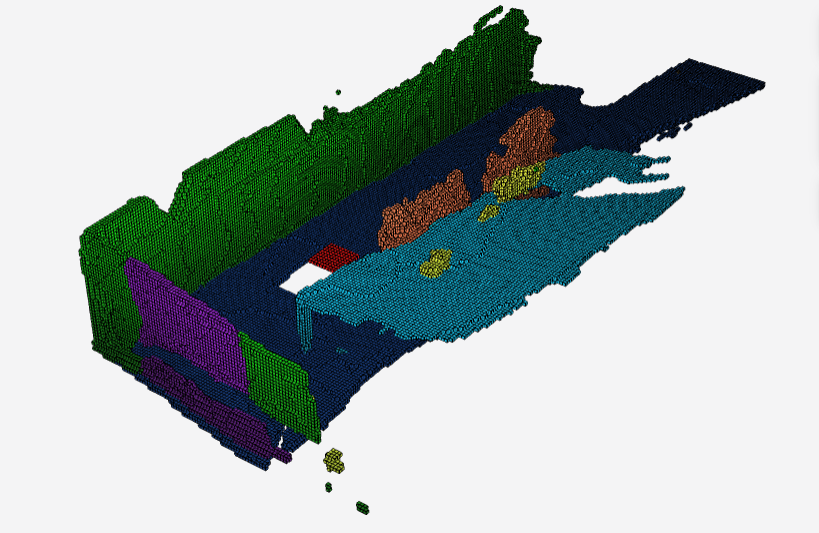}
    \end{subfigure}
    \begin{subfigure}[b]{0.15\textwidth}
        \centering
        \includegraphics[width=\textwidth, height=0.8\textwidth]{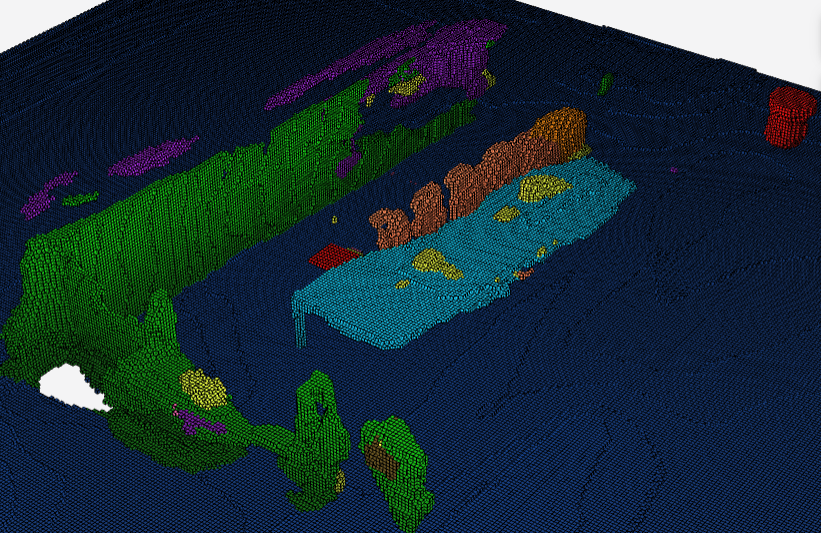}
    \end{subfigure}
    \begin{subfigure}[b]{0.15\textwidth}
        \centering
        \includegraphics[width=\textwidth, height=0.8\textwidth]{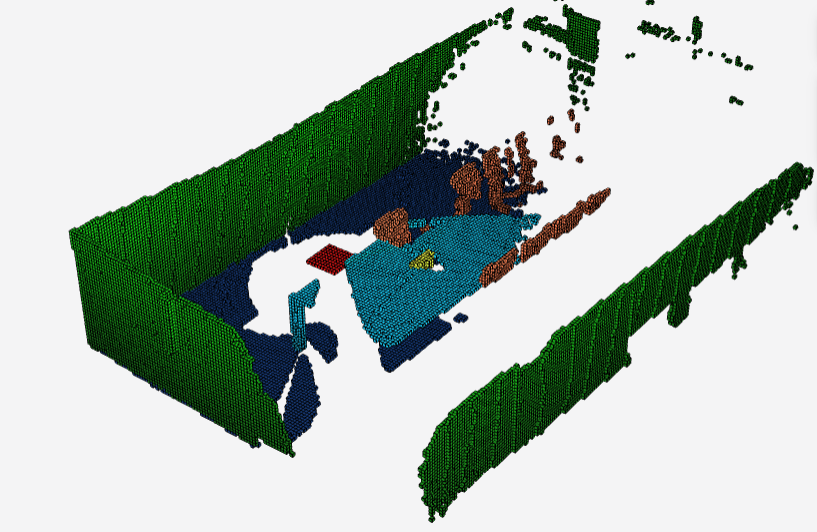}
    \end{subfigure}

    \vspace{1mm}
    \begin{adjustbox}{rotate=90}\hspace{2mm}Appartment\end{adjustbox}
    \begin{subfigure}[b]{0.15\textwidth}
        \centering
        \begin{minipage}[b][0.8\textwidth][t]{0.8\textwidth}
            \includegraphics[width=0.5\textwidth,height=0.5\textwidth]{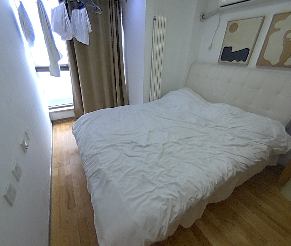}%
            \includegraphics[width=0.5\textwidth,height=0.5\textwidth]{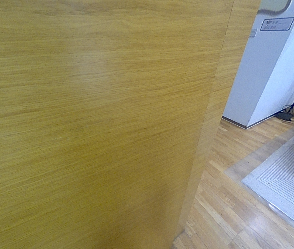}\\[-0.5mm]
            \includegraphics[width=0.5\textwidth,height=0.5\textwidth]{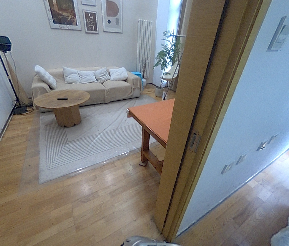}%
            \includegraphics[width=0.5\textwidth,height=0.5\textwidth]{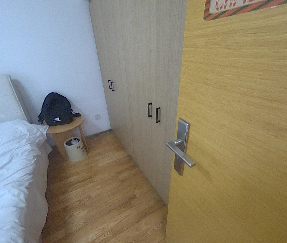}
        \end{minipage}
        \caption*{Images}
    \end{subfigure}\hspace{-2mm}
    \begin{subfigure}[b]{0.15\textwidth}
        \centering
        \includegraphics[width=\textwidth, height=0.8\textwidth]{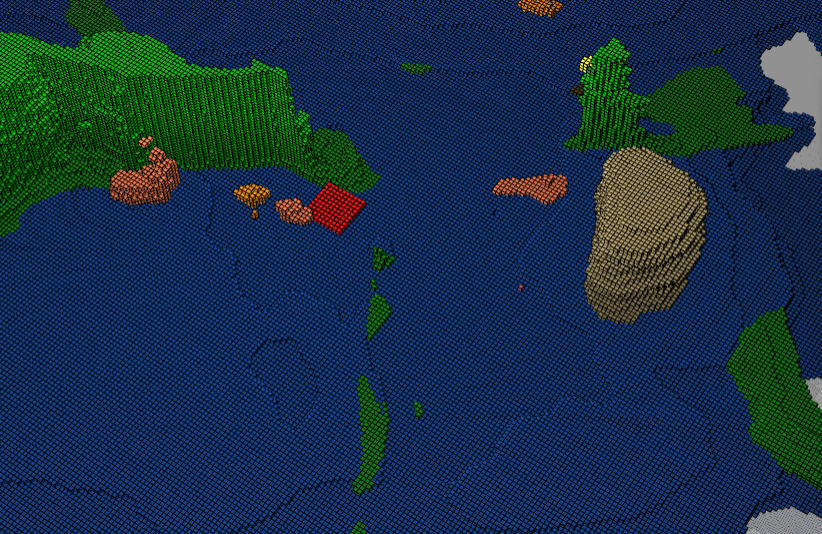}
        \caption*{Flash-Occ}
    \end{subfigure}
    \begin{subfigure}[b]{0.15\textwidth}
        \centering
        \includegraphics[width=\textwidth, height=0.8\textwidth]{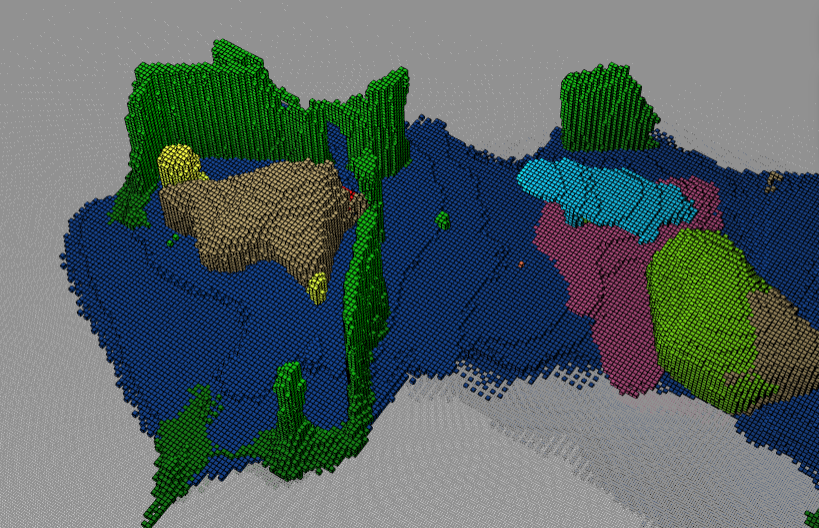}
        \caption*{FB-Occ}
    \end{subfigure}
    \begin{subfigure}[b]{0.15\textwidth}
        \centering
        \includegraphics[width=\textwidth, height=0.8\textwidth]{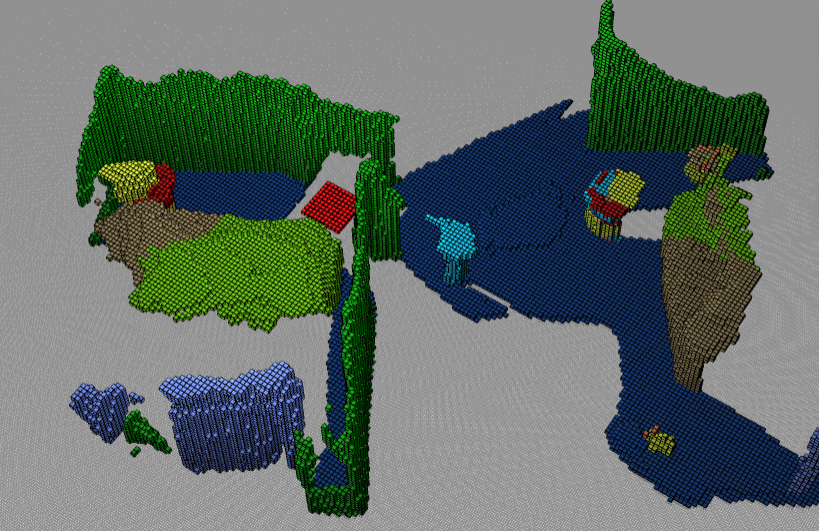}
        \caption*{SurrOcc}
    \end{subfigure}
    \begin{subfigure}[b]{0.15\textwidth}
        \centering
        \includegraphics[width=\textwidth, height=0.8\textwidth]{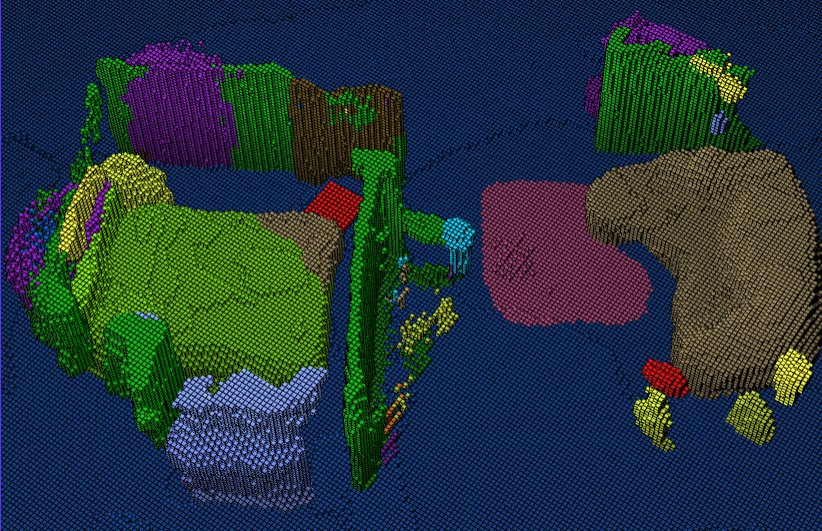}
        \caption*{\method (Ours)}
    \end{subfigure}
    \begin{subfigure}[b]{0.15\textwidth}
        \centering
        \includegraphics[width=\textwidth, height=0.8\textwidth]{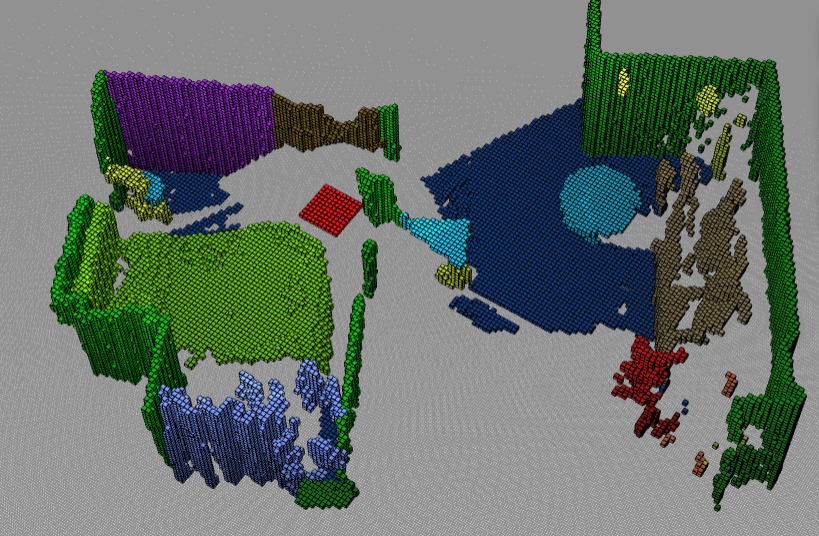}
        \caption*{GT}
    \end{subfigure}
    
    \caption{\textbf{Qualitative comparisons on the test set (first row) and real-world scenes (rows 2--3).} The first column shows the left images from the four stereo pairs (\textit{front}, \textit{back}, \textit{left}, and \textit{right}).}
    \label{fig:qualitative}
\end{figure}
\subsection{Qualitative Comparisons}

As shown in Figure~\ref{fig:qualitative}, our stereo-based model (\textbf{\method}) generates more complete and geometrically consistent occupancy reconstructions than the monocular baseline Flash-Occ, FB-Occ, and SurroundOcc.
Across both test scenes and real-world environments, \textbf{\method} preserves stable geometric structures and clear spatial layouts, highlighting its strong robustness and superior sim-to-real generalization capability.

\section{Conclusion}

In this work, we presented Humanoid-OmniOcc, a panoramic stereo-based occupancy benchmark for embodied humanoid perception, featuring high-quality voxel-level annotations across diverse simulated and real-world indoor environments. By aligning simulation with real sensor configurations under a Real2Sim2Real paradigm, the benchmark enables scalable training while preserving strong sim-to-real consistency.
Based on this dataset, we introduced \method{}, a stereo-guided occupancy network that exploits reliable depth priors for improved 2D-to-3D lifting. Experimental results show that stereo cues significantly enhance geometric fidelity and generalization ability over monocular baselines, especially under real-world domain shifts.
Overall, our results highlight the importance of stereo-aware perception and realistic simulation design for embodied 3D understanding. We hope Humanoid-OmniOcc can serve as a foundation for future research in omnidirectional perception, humanoid navigation, and interactive embodied intelligence.

\bibliographystyle{unsrtnat}  
\bibliography{main}

\newpage
\appendix

\section*{Appendix}

\subsection{Overview of the Humanoid-OmniOcc Dataset}
As summarized in Table~\ref{tab:datasample}, \textbf{Humanoid-OmniOcc} dataset comprises a total of 155,753 samples,
each containing four stereo pairs captured from omnidirectional viewpoints.
The data are collected from \textbf{fifteen} simulated indoor environments, each featuring distinct spatial layouts and aesthetic styles, covering diverse room types including apartments, studios, bedrooms, dining rooms, kitchens, lounges, living rooms, offices, and patios.
In addition, five real-world indoor environments (Bar, Corridor, Office, and Apartment) are captured to enable sim-to-real evaluation.
The entire dataset construction and physically-based rendering process consumed approximately \textbf{4,000 GPU hours}
on NVIDIA L20 clusters, encompassing global illumination simulation, stereo rendering,
and voxel-level occupancy ground-truth generation.



\begin{wraptable}{r}{0.48\textwidth}
    \centering
    \vspace{-10pt}
    \caption{\textbf{Dataset samples distribution.}}
    \label{tab:datasample}
    \setlength{\tabcolsep}{4pt}
    \renewcommand{\arraystretch}{1.12}
    \resizebox{\linewidth}{!}{%
    \begin{tabular}{l r | l r}
        \toprule
        \rowcolor{gray!20}
        \textbf{Scene} & \textbf{Frames}
        & \textbf{Scene} & \textbf{Frames} \\
        \midrule
        ApartM1  & 10,766 & Abode   & 4,951  \\
        Studio   & 14,860 & Loft    & 15,694 \\
        BedR1    & 8,912  & Nooks   & 19,851 \\
        Patio    & 5,705  & BedR2   & 5,300  \\
        DiningR  & 9,909  & Kitchen & 6,116  \\
        Lounge   & 4,706  & LivingR & 10,206 \\
        ApartM2  & 10,860 & Work    & 13,351 \\
        Office   & 14,566 & Sum     & 155,753 \\
        \bottomrule
    \end{tabular}
    }
    \vspace{-10pt}
\end{wraptable}

\subsection{Ground Truth Generation in Simulation}
At each frame, Isaac Sim provides RGB-D images for all stereo cameras, the robot's pose $T_{w}^{r}$ in the world coordinate system, and each camera's pose $T_{w}^{c_i}$. The local environment within $[\text{bev}_w, \text{bev}_d, \text{bev}_h]$ is exported as a mesh and voxelized with a voxel size of $\text{voxel\_size}$. For each camera, the extrinsic matrix is computed as $\text{Extrinsic}_{c_i} = (T_{w}^{c_i})^{-1} T_{w}^{r}$. Occupancy voxels in the world coordinates $P_w$ are transformed into the robot base frame $P_b = (T_{w}^{r})^{-1} P_w$, and further into each camera coordinate frame $P_c = \text{Extrinsic}_{c_i} P_b$. 
Valid points are filtered by $P_c(z) > 0.15$ and projected to the image plane using intrinsic parameters to obtain pixel coordinates and depth $d_{pt}$. Each voxel is labeled following the depth consistency rule:
\begin{equation}
\text{Label}(v) =
\begin{cases}
0~(\text{free}), & \text{if } d_v < d_{\text{img}} - \Delta_v, \\[6pt]
255~(\text{unknown}), & \text{if } d_v > d_{\text{img}} + \Delta_v, \\[6pt]
1~(\text{occupied}), & \text{if } |d_v - d_{\text{img}}| \le 2\Delta_v,
\end{cases}
\label{eq:label}
\end{equation}
All voxel grids are projected back to the image plane for visual inspection. The projection visualization allows intuitive verification of occupancy accuracy and alignment consistency.

To support fine-grained scene understanding, we annotate every occupied voxel 
with one of 15 semantic categories: \textit{free, floor, partitions} (wall, 
railings, window), \textit{door, chair, table, sofa, bed, appliance} (large 
household items), \textit{cabinet, carpet, plant, objects} (miscellaneous 
small items), \textit{suspended objects} (curtains, hanging fixtures), and 
\textit{other}. Categories are defined by merging simulator object labels 
into coherent groups that align with common indoor object taxonomies. 
The label distribution across scenes shows balanced coverage of structural 
(\textit{floor, partitions, door}) and furniture (\textit{chair, table, 
sofa, bed}) categories.

\subsection{Ground Truth Generation in the Real World}
We generate 3D occupancy supervision ground truth based on LiDAR point clouds. First, the fused dense point clouds undergo a preprocessing stage to eliminate invalid points, extreme outliers, and significant noise, thereby mitigating the impact of artifacts on voxel labeling. Subsequently, the point clouds are transformed into a unified baselink coordinate system using sensor calibration parameters, and a predefined 3D voxel space is constructed around the robotic. Each point coordinate is discretized into its corresponding voxel unit, where voxels containing LiDAR points are labeled as occupied, representing the regions supported by actual physical observations.
To prevent unobserved or occluded regions from being erroneously categorized as free space, we further incorporate a visibility assessment based on viewing frustum constraints. Specifically, camera intrinsic and extrinsic parameters are utilized to determine whether each voxel resides within the effective field of view (FoV) of the surround-view cameras. For voxels within the FoV that lack LiDAR point occupancy, we employ the 3D Bresenham ray-tracing algorithm to cast rays from the sensor center toward the target voxels. By traversing the voxels along the ray path and checking for the presence of occupied voxels, we determine the visibility status: if a ray is intercepted by an occupied voxel before reaching its destination, the target voxel is considered to be in an occluded region—meaning its true state cannot be resolved by current observations—and is thus marked as unknown; otherwise, it is classified as free space. Through this pipeline, the generated occupancy ground truth simultaneously characterizes the occupied space directly observed by LiDAR, the free space within the visible range, and the unknown space resulting from occlusions, providing a more physically consistent 3D spatial supervision signal for model training.
Finally, semantic labels are manually annotated to obtain semantic occupancy supervision.

\begin{figure}[htbp]
    \centering
    \begin{subfigure}[t]{0.48\linewidth}
        \centering
        \includegraphics[width=\linewidth]{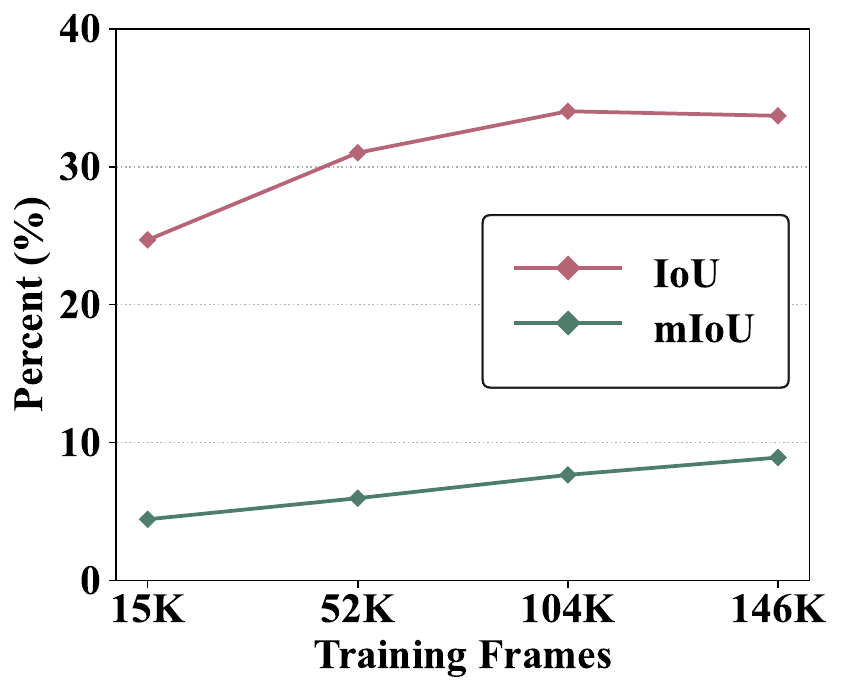}
        \label{fig:training_frames}
    \end{subfigure}
    \hfill
    \begin{subfigure}[t]{0.48\linewidth}
        \centering
        \includegraphics[width=\linewidth]{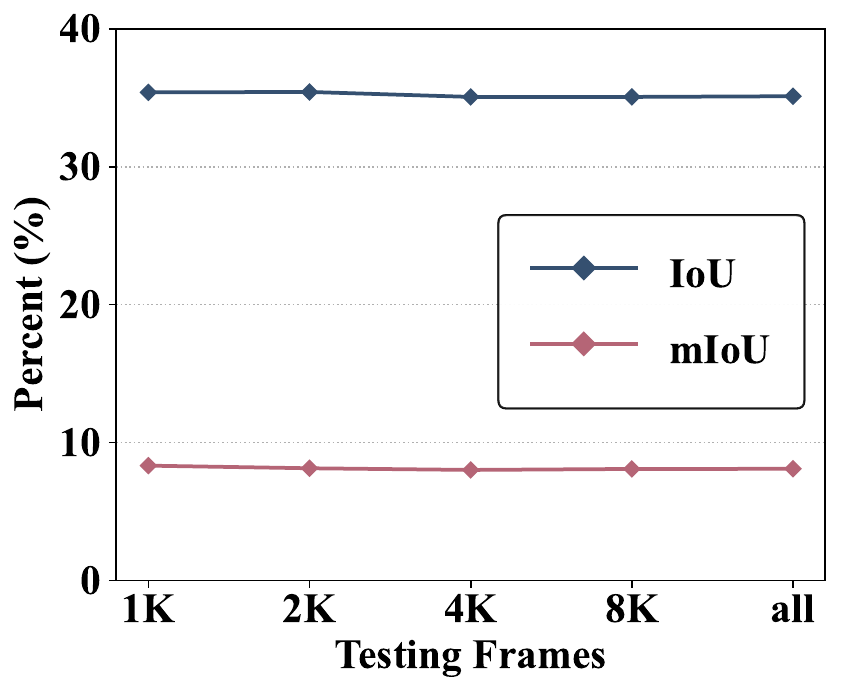}
        \label{fig:testing_frames}
    \end{subfigure}
    \caption{\textbf{Performance scaling of \method~with training and testing data.}}
    \label{fig:data_scaling}
\end{figure}

\subsection{More Analysis on the Humanoid-OmniOcc Dataset}

Figure~\ref{fig:data_scaling} (Left) illustrates the influence of training data scale on occupancy prediction performance. 
As the number of training frames increases from 15K to 59K, all three metrics---IoU, Precision, and Recall---consistently improve, indicating that larger scene coverage leads to better spatial generalization and completeness. 
We also analyze the robustness of \method~under varying testing frame scales, as shown in Figure~\ref{fig:data_scaling} (Right). 
When the number of testing frames increases from 1K to 9K, all three metrics---IoU, Precision, and Recall---remain nearly constant, with fluctuations within only 1--2\%. 
This consistency demonstrates that the learned representation generalizes well across different test scenes, confirming that the model does not rely on scene-specific priors or temporal bias.


\end{document}